%% file: arxivversion.tex
\documentclass{article}

\usepackage{arxiv}

\usepackage[utf8]{inputenc} 
\usepackage[T1]{fontenc}    
\usepackage{hyperref}       
\usepackage{url}            
\usepackage{booktabs}       
\usepackage{amsfonts}       
\usepackage{nicefrac}       
\usepackage{microtype}      
\usepackage[pdftex]{graphicx}
\usepackage{amssymb,amsfonts,amsmath,amscd}
\usepackage{amscd}
\usepackage[printonlyused,withpage]{acronym}
\usepackage{natbib}
\usepackage{mathrsfs}
\usepackage{breqn}
\usepackage{bm}
\usepackage{tabularx,booktabs}
\newcommand{\jobcolor}{}

\usepackage{cuted}
\graphicspath{{figs/}}

\acrodef{BA}{Bundle Adjustment}
\acrodef{DNN}{Deep Neural Network}
\acrodef{EM}{Expectation Maximization}
\acrodef{GEM}{Generalized Expectation Maximization}
\acrodef{HMM}{Hidden Markov Model}
\acrodef{LDS}{Linear Dynamical System}
\acrodef{LQG}{Linear Quadratic Gaussian}
\acrodef{LQR}{Linear Quadratic Regulator}
\acrodef{LTI}{Linear Time-Invariant}
\acrodef{RTS}{Rauch-Tung-Striebel}
\acrodef{SGD}{Stochastic Gradient Descent}
\acrodef{SLAM}{Simultaneous Localization and Mapping}
\acrodef{RKHS}{Reproducing Kernel Hilbert Space}
\acrodef{SMW}{Sherman-Morrison-Woodbury}

\acrodef{GVI}{Gaussian Variational Inference}
\acrodef{ESGVI}{Exactly Sparse Gaussian Variational Inference}
\acrodef{MAP}{Maximum A Posteriori}
\acrodef{ML}{Maximum Likelihood}
\acrodef{KL}{Kullback-Leibler}
\acrodef{PDF}{Probability Density Function}
\acrodef{NEES}{Normalized Estimation Squared Error}
\acrodef{KF}{Kalman Filter}
\acrodef{VKF}{Variational Kalman Filter}
\acrodef{ISPKF}{Iterated Sigmapoint Kalman Filter}
\acrodef{ESGVI-GN}{ESGVI Gauss-Newton}
\acrodef{ELBO}{Evidence Lower Bound}
\acrodef{NGD}{Natural Gradient Descent}
\acrodef{FIM}{Fisher Information Matrix}
\acrodef{RANSAC}{Random Sample And Consensus}
\acrodef{IRLS}{Iteratively Reweighted Least-Squares}
\acrodef{BRD}{Black-Rangarajan Duality}
\acrodef{GNC}{Graduated Non-Convexity}

\hypersetup{%
    pdftitle={},
    pdfauthor={},
    pdfkeywords={},
    pdfsubject={},
    pdfstartview=FitH,%
    bookmarks=true,%
    bookmarksopen=true,%
    breaklinks=true,%
    colorlinks=true,%
    linkcolor=black,
    anchorcolor=black,%
    citecolor=black,
    filecolor=black,%
    menucolor=black,
    pagecolor=black,%
    urlcolor=black
}%

\def\so{}

\input{notation_header}

\input{def.tex}

\begin{document}

\title{On the Eigenstructure of Rotations and Poses:
Commonalities and Peculiarities}

\date{}

\author{\large Gabriele M. T. D'Eleuterio and Timothy D. Barfoot\\[3pt]
\normalsize University of Toronto}

\maketitle


\begin{abstract}
Rotations and poses are ubiquitous throughout many fields of science and engineering such as robotics, aerospace, computer vision and graphics.  In this paper, we provide a complete characterization of rotations and poses in terms of the eigenstructure of their matrix Lie group representations, ${SO(3)}$, ${SE(3)}$ and ${\Ad(SE(3))}$.  An eigendecomposition of the pose representations reveals that they can be cast into a form very similar to that of rotations although the structure of the former can vary depending on the relative nature of the translation and rotation involved.  Understanding the eigenstructure of these important quantities has merit in and of itself but it is also essential to appreciating such practical results as the minimal polynomial for rotations and poses and the calculation of Jacobians; moreover, we can speak of a principal-axis pose in much the same manner that we can of a principal-axis rotation.
\end{abstract}

\keywords{
rotations,  poses,  eigenanalysis, matrix Lie groups,  Lie algebras
}

\section{Introduction}

The conjoined representation of position and orientation---commonly termed {\em pose}---is of central importance to nearly all studies of vehicular motion and more generally rigid-body motion in such applications as robotics (including and perhaps especially robotic manipulators), aerospace, computer vision and graphics.  Position can be represented by vectors in three-dimensional space; orientation (attitude in aerospace applications) can be dressed in a variety of guises, e.g., Euler angles, Euler parameters (quaternions), axis-angle variables.  The rotational transformation matrix is the iconic expression of the orientation of one reference frame with respect to another or of a body with respect to some frame.  Unlike position vectors, rotation matrices (or any other incarnation of orientation) do not commute.  The loss of this property naturally extends to pose representations.

To offer some consolation, however, rotation matrices form a {\em Lie group}, the {\em special orthogonal group} $SO(3)$, and as a Lie group there is a corresponding {\em Lie algebra}, wherein the rotations or orientations can be identified with vectors in a nonassociative algebra.  The same is true of pose representations.  Two common representations of pose prevail.  One, the {\em special Euclidean group} $SE(3)$, which is widely used in kinematics \citep{Belta&Kumar:2002,Grassmann&Burgner-Kahrs:2019}, robotics \citep{barfoot_tro14,Huber&Wollherr:2020} and computer vision \citep{Tron&al:2008,Arrigoni&al:2018}, uses $4\times4$ matrices \citep{denavit&hartenberg:1955, Murray&al:1994}.  The other is the pose-adjoint representation, denoted $\Ad(SE(3))$ as a group, uses $6\times6$ matrices and is typically preferred for dynamics \citep{Featherstone:1987,Hughes&Sincarsin:1989,D'Eleuterio:1992,Sincarsin&al:1993,Park&al:1995}.  These latter two groups, as they both combine translation and rotation, are closely related.  In appearance and function, though, $\Ad(SE(3))$ more closely resembles $SO(3)$; see also the fine exposition by \citet{Borri&al:2000}.

Now, in the case of rotation, the vectors of the accompanying Lie algebra $\so(3)$ are described by the rotation axis and angle of Euler's theorem.  In the case of pose, they are given by the displacement along with the rotation axis and angle as in the Mozzi-Chasles theorem \citep{mozzi1763,chasles1830}.  The former of course define a three-dimensional space while the latter, in either $\se(3)$ or $\ad(\se(3))$, establish a six-dimensional space.

The elements of these Lie groups may be expressed as an exponential map of a vector in its Lie algebra.  For example, as is well known, a rotation matrix can be canonically expressed as an exponential of the skew-symmetric matrix representing the axis-angle vector.  Similar expressions hold for $SE(3)$ and $\Ad(SE(3))$ \citep{Park&al:1995,barfoot_ser17}.

Given the intimate relationship between a Lie group and its Lie algebra, it is not surprising that rotation and pose representations can be completely characterized in terms of the eigenstructure of their affiliated Lie-algebraic elements.  However, pose representations can be plagued by repeated eigenvalues that do not admit a full set of linearly independent eigenvectors.  Hence, these representations require a Jordan, or more generally a block upper-triangular, decomposition.

In this paper, we present a detailed eigenanalysis of these three important Lie groups---$SO(3)$, $SE(3)$ and $\Ad(SE(3))$---that pervade many fields of science and engineering.  We shall alternately do so in complex terms, for eigenvalues and eigenvectors are complex, and in real terms, where the decomposition of the group elements lead to {\it principal-axis rotations} or {\it principal-axis poses}.  We shall remark on the similarities among the groups as well as highlight some of their peculiarities.  Our aim is to provide an interconnected understanding of rotation and pose representations by examining their eigenstructures.  While an eigenanalysis of pose can be exploited to identify the Mozzi-Chasles axis, see in particular the notable work of Bauchau and colleagues \citep{Bauchau&Choi:2003,Bauchau&Li:2011,Bauchau:2011}, a thorough exposition on the eigenstructure of pose has not to our knowledge been previously reported in the literature.

\section{Rotations and Poses as Matrix Lie Groups}

Let us begin with a brief introduction to the Lie groups $SO(3), SE(3)$ and $\Ad(SE(3))$.  The first of these is the {\em special orthogonal group} in three dimensions:
\begin{equation}
\label{eq:SO3}
SO(3) = \left\{  \mbf{C} \in \mathbb{R}^{3\times3} \; | \; \mbf{C}^T \mbf{C} = \mbf{1}, \mbox{det} \,\mbf{C} = 1 \right\}.
\end{equation}
This is the group of familiar rotation matrices.  The second is the {\em special Euclidean group} in three dimensions:
\begin{equation}
\label{eq:se3}
SE(3) = \left\{  \mbf{T} = \bbm \mbf{C} & \mbf{r} \\ \;\,\mbf{0}^T & 1 \ebm \in \mathbb{R}^{4\times4} \; \Biggl| \; \mbf{C} \in SO(3), \, \mbf{r} \in \mathbb{R}^3 \right\}.
\end{equation}
Its elements represent pose, i.e., rotation in $\mbf{C}$ and translation in $\mbf{r}$.  An alternate representation of pose is given by
\begin{equation}
\bsT = \bbm \mbf{C} & \mbf{r}^\times \mbf{C} \\ \mbf{O} & \mbf{C} \ebm
\end{equation}
where
\begin{equation}\label{cross:1}
  \mbf r^\times = \bbm 0 & -r_3 & r_2 \\ r_3 & 0 & -r_1 \\ -r_2 & r_1 & 0 \ebm
\end{equation}
for $\mbf r = [\, r_1 \;\; r_2 \;\; r_3\,]^T$.  (The matrix ``cross'' operator mimics the cross-product operator for vectors.)  Accordingly, we define
\begin{equation}\label{eq:se3adjointmap1}
\Ad(SE(3))=
 \left\{  \Tbig = \bbm \mbf{C} & \mbf{r}^\times \mbf{C} \\ \mbf{0} & \mbf{C} \ebm \in \mathbb{R}^{6\times6} \; \Biggl| \; \mbf{C} \in SO(3), \, \mbf{r} \in \mathbb{R}^3 \right\},
\end{equation}
as the group of the {\em pose adjoints}.

In each case, the group elements may be viewed in two ways.  They may be used to represent the rotation or pose of a rigid body but they may also be viewed as transformations for vector quantities from one reference frame or reference system to another.  (We regard a reference frame as defining a directional basis without regard for a specific origin; a reference system $(O_a,\scF_a)$ implies a reference frame $\scF_a$ plus a reference point or origin $O_a$.)

In $SO(3)$, $\mbf{C}$ transforms the coordinates of any vector from one dextral orthogonal reference frame to another.  With $SE(3)$, $\mbf{T}$ transforms quantities of the form
\begin{equation}\label{homo:1}
  \mbf s = \bbm \mbf{t} \\ 1 \ebm \in \Re^4,
\end{equation}
where $\mbf t$ is the position of an arbitrary point $P$ relative to one reference system (i.e., relative to a reference point of one frame and expressed in that frame) to the position of $P$ relative to another reference system; the fourth entry is the number 1.  The column $\mbf s$ in (\ref{homo:1}) constitutes the {\em homogeneous coordinates} of $P$ and $\mbf T$ is the {\em homogeneous representation} of the transformation \citep{Murray&al:1994}.  This type of transformation is used extensively, for example, in the analysis of mechanisms and robotics as part of the Denavit-Hartenberg convention \citep{denavit&hartenberg:1955}.

The $\Ad(SE(3))$ group offers transformations from one reference system to another for quantities of the form
\begin{equation}
  \mbs{t} = \bbm \mbf{t} \\ \mbs{\tau} \ebm \in \Re^6,
\end{equation}
where $\mbf{t}$ is a translational quantity and $\mbs{\tau}$ is a rotational quantity and one or the other is associated with the reference point; that is, one of $\mbf t$ or $\mbs\tau$ is a {\it bound vector}, respecting a reference point, and the other is a {\it free vector}.  For example, the generalized velocity or twist velocity of a rigid body, in which $\mbf{t}$, the bound vector, is the velocity of a point affixed to the body and $\mbs{\tau}$, the free vector, is its angular velocity relative to some established frame, the transformation is given by $\bsT\mbs{t}$.  More specifically, if $\mbs t_a$ is the generalized velocity relative to $(O_a,\scF_a)$ then $\mbs t_b = \bsT_{ba}\mbs t_a$ is the generalized velocity relative to $(O_b,\scF_b)$.  Infinitesimal twist (infinitesimal rotation and translation) is transformed by $\bsT$ as well.

On the other hand, for the generalized momentum of a rigid body, it is the momentum of the body $\mbf{t}$ that is the free vector and the angular momentum $\mbs{\tau}$ about a given point in the body is the bound vector.  In this case, the transformation is instead $\bsT^T\mbs{t}$, i.e., $\mbs t_a = \bsT_{ba}^T\mbs t_b$ \citep{Hughes&Sincarsin:1989,Bauchau&Choi:2003}.  Note that the transformation here is from $(O_b,\scF_b)$ to $(O_a,\scF_a)$.   Generalized force or wrench, in which $\mbf{t}$ is force and $\mbs{\tau}$ is torque about a given point, also transforms in this manner.

\subsection{Group-Theoretic Concepts}

It will prove worthwhile to take a brief excursion into Lie group theory.    Each of the above Lie groups has associated with it a Lie algebra, the tangent space of the group at its identity element.  The {\em Lie bracket} $[\cdot,\cdot]$, which characterizes a Lie algebra, is a bilinear operation, mapping back to the algebra, satisfying alternativity,
$[u,u] = 0$, and Jacobi's identity, $[u,[v,w]] + [v,[w,u]] + [w,[u,v]] = 0$, for all $u,v,w$ in the algebra.  For matrix Lie algebras, the Lie bracket is the commutator of matrices, $[u,v] = uv - vu$.

For $SO(3)$, the elements of its Lie algebra $\so(3)$ are the $3\times3$ skew-symmetric matrices $\bfom^\times$.  Note that $\so(3)$ is isomorphic to $\Re^3$ as elements can be represented simply by $\bfom$.  The Lie bracket of $\so(3)$ is $[\bfom^\times,\bfv^\times] = \bfom^\times\bfv^\times - \bfv^\times\bfom^\times$; however, in $\Re^3$ it can be expressed as $\bfom^\times\bfv$, the standard cross product---note that the Lie bracket is identically given by $[\bfom^\times,\bfv^\times] = (\bfom^\times\bfv)^\times$.

For $SE(3)$, the Lie algebra is denoted $\se(3)$ and the elements are
\begin{equation}\label{adj:1}
  \bfup^\wdg = \bbm \bfom^\times & \bfv \\ \0^T & 0 \ebm \in \Re^{4\times4},
\end{equation}
where
\begin{equation}
  \bfup = \bbm \bfv \\ \bfom \ebm \in \Re^6.
\end{equation}
In like fashion to $\so(3)$, $\se(3)$ is isomorphic to $\Re^6$.  The Lie bracket here is $[\bfup^\wdg,\bfch^\wdg] = \bfup^\wdg\bfch^\wdg - \bfch^\wdg\bfup^\wdg$.  This product is an element in $\se(3)$; however, demonstrating this is most easily done by first defining a ``cross product'' in $\Re^6$.

Define
\begin{equation}\label{cross:2}
  \bfup^\Wdg = \bbm \bfom^\times & \bfv^\times \\ \bfO & \bfom^\times \ebm \in \Re^{6\times6}.
\end{equation}
It is straightforward to prove $(\bfup^\Wdg\bfch)^\wdg = [\bfup^\wdg,\bfch^\wdg]$, which shows that indeed $[\bfup^\wdg,\bfch^\wdg]$ is an element in $\se(3)$.  (The definitions (\ref{cross:1}) and (\ref{cross:2}) for the cross operation are distinguished by their operand.)

For every $\bfT \in SE(3)$, we can define the {\em adjoint map} $\Ad_\bfT:\se(3)\rightarrow\se(3)$ such that
\begin{equation}\label{adj:1a}
  \Ad_\bfT\bfup^\wdg = \bfT\bfup^\wdg\bfT^\inv.
\end{equation}
The result of this operation is
\begin{equation}\label{adT:1}
  \!\Ad_\bfT\bfup^\wdg = \bbm (\bfC\bfom)^\times & \bfr^\times\bfC\bfom + \bfC\bfv \\
    \0^T & 0 \ebm = \bbm \bfr^\times\bfC\bfom + \bfC\bfv \\ \bfC\bfom \ebm^\wdg
\end{equation}
upon noting the identity that $(\bfC\bfom)^\times = \bfC\bfom^\times\bfC^T$.  The last quantity in (\ref{adT:1}), unwedged, is in fact just $\bsT\bfup$, which suggests that the adjoint map transforms $\bfup^\wdg$ by $\bfT$ or, equivalently, $\bfup$ by $\bsT$.  This has led to a slightly corrupting notation by writing $\Ad_\bfT = \bsT$, e.g., see \citet{Park&al:1995}.  We also note that $\Ad^\ast_\bfT = \bsT^T$ is the dual transformation; if $\bfup$ is transformed by $\bsT$ then $\bfup^\wdg \in \se(3)$ whereas if $\bfup$ is transformed by $\bsT^T$ then ${\bfup^\wdg}^T \in \se^\ast(3)$, the dual of $\se(3)$.  This helps explain the nature of the different transformations for generalized momentum and wrench vis-\`a-vis twist velocity.  This also explains the terminology of ``pose adjoints'' for the $6\times6$ pose representation and the use of the
notation $\Ad(SE(3))$ for the corresponding Lie group.

We mention for completeness that the {\em adjoint representation}
for $\bfup^\wdg \in \se(3)$ is the map $\ad_{\bfup^\wdg}:\se(3) \rightarrow \se(3)$ where
\begin{equation}
  \ad_{\bfup^\wdg}\bfch^\wdg = [\bfup^\wdg,\bfch^\wdg].
\end{equation}
As noted, $\ad_{\bfup^\wdg}\bfch^\wdg = [\bfup^\wdg,\bfch^\wdg] = (\bfup^\Wdg\bfch)^\wdg$ suggesting again, with a similar
abuse of notation, that $\ad_{\bfup^\wdg} = \bfup^\Wdg$.  Hence, $\ad(\se(3))$ denotes the Lie algebra of $\Ad(SE(3))$ and its elements are $\bfup^\Wdg$.  As with $\se(3)$, $\ad(\se(3))$ is isomorphic to $\Re^6$ and the operation $\bfup^\Wdg\bfch$ corresponds to the commutator of $\ad(\se(3))$---note that $[\bfup^\Wdg,\bfch^\Wdg] = \bfup^\Wdg\bfch^\Wdg - \bfch^\Wdg\bfup^\Wdg = (\bfup^\Wdg\bfch)^\Wdg$---just as the operation $\bfom^\times\bfv$ corresponds to the commutator of $\so(3)$.

\begin{table*}[t!]
 \caption{Identities in $SO(3)$ and $\mbox{\normalfont Ad}(SE(3))$}
\label{table}
\begin{center}
\begin{tabular}{ccc}
\toprule
$SO(3)$ & & $\mbox{Ad}(SE(3))$ \\
\hline
\addlinespace
    $\bfu^\times\bfu = \0$ & $\longleftrightarrow$ &  $\bfxi^\Wdg\bfxi = \0$ \\
    $\bfu^\times\bfv = -\bfv^\times\bfu$  & $\longleftrightarrow$ & $\bfxi^\Wdg\bfet = -\bfet^\Wdg\bfxi$ \\
    $(\bfu^\times\bfv)^\times = \bfu^\times\bfv^\times - \bfv^\times\bfu^\times$  & $\longleftrightarrow$ &
    $(\bfxi^\Wdg\bfet)^\Wdg = \bfxi^\Wdg\bfet^\Wdg - \bfet^\Wdg\bfxi^\Wdg$ \\
    $\bfu^\times\bfv^\times\bfw + \bfv^\times\bfw^\times\bfu + \bfw^\times\bfu^\times\bfv = \0$
    & $\longleftrightarrow$ & $\bfxi^\Wdg\bfet^\Wdg\bfze + \bfet^\Wdg\bfze^\Wdg\bfxi + \bfze^\Wdg\bfxi^\Wdg\bfet = \0$ \\
\addlinespace
\bottomrule
\end{tabular}
\end{center}
\end{table*}

\paragraph{Some Identities.}
There is a particular affinity between $\so(3)$ and $\ad(\se(3))$.  Consider some identities given in Table~\ref{table}\ that translate directly from $\so(3)$ to $\ad(\se(3))$.
The first and last (Jacobi's identity) of these establish the commutators in $\Re^3$ and $\Re^6$, respectively, as Lie brackets.  The second identity expresses skew-commutativity, which is per force a consequence of the Lie bracket and the third shows that the product of the Lie bracket is an element in the Lie algebra.

There are some identities involving the rotation and pose representations that are also worthy of note \citep{barfoot_ser17}:
\begin{equation}\label{pose:8}
    (\bfC\mbf u)^\times = \bfC\mbf u^\times\bfC^T,\qquad
    (\bsT\mbs\xi)^\wdg = \mbf T\mbs\xi^\wdg\mbf T^{-1},\qquad
    (\bsT\mbs\xi)^\Wdg = \Tbig\mbs\xi^\Wdg\Tbig^{-1},
\end{equation}
where $\mbf u \in \Re^3$ and $\mbs\xi \in \Re^6$.  The second identity is essentially (\ref{adj:1a}).

\subsection{Kinematics of Rotations and Poses}

There exists an exponential map from each element in the Lie algebra to an element in its corresponding Lie group \citep{Marsden&Ratiu:2002, ivanovic}; that is, every rotation and pose matrix representation can be written as the exponential of some matrix.  We can better appreciate this mapping by considering the kinematics of rotations and poses.

To begin, Poisson's kinematical relation gives the time evolution of $\mbf C \in SO(3)$, namely,
\begin{equation}\label{poisson:1}
  \dot{\mbf C} = \mbs\omega^\times\mbf C.
\end{equation}
(We point out that Poisson's equation is sometimes written with a minus sign depending on the definition of $\mbs\omega$.  Practitioners of robotics tend to prefer it positively expressed; those in aerospace usually opt for the negative sign.  We, despite our aerospace background, will forego the minus as the exponential map of $\mbs\omega^\times$ will figure heavily here and the absence of the sign will make the notation slightly more streamlined.)  We should furthermore clarify that if $\bfC = \bfC_{ba}$ is the rotation matrix from $\scF_a$ to $\scF_b$ then $\mbs\om = \mbs\om_b^{ab}$ is the angular velocity of $\scF_a$ relative to $\scF_b$ as expressed in $\scF_b$.

According to Euler's theorem any sequence of rotations can be reduced to a single rotation about a single axis.  We would then be permitted, with loss in generality, to consider $\mbs\omega$ to
be constant in achieving any desired rotation.  As such, the rotation matrix is given by
\begin{equation}\label{rot:1}
  \mbf C = \exp(\mbs\omega^\times t).
\end{equation}
This in fact expresses a fundamental result in Lie groups that there exists a unique curve that maps the line $\mbs\omega^\times t$ in the Lie algebra onto a one-parameter subgroup of the corresponding Lie group.

We may furthermore write
\begin{equation}\label{rot:2}
  \mbf C(\mbs\phi) = \exp\mbs\phi^\times
\end{equation}
by identifying $\mbs\phi = \mbs\omega t = \phi\mbf a$, in which $\mbf a$ is the (normalized) Euler axis and $\phi$ the angle of rotation at time $t$.  (We can replace $t$ with any other surrogate variable on which $\mbf C$ may depend.)  The skew-symmetric matrix $\mbs\phi^\times$ is an element in $\so(3)$ and (\ref{rot:2}) is the exponential map for rotation matrices.

Turning our attention to $\mbf T \in SE(3)$, we can take the derivative to reveal that
\begin{equation}\label{pose:1}
  \dot{\mbf T} = \bbm \dot{\mbf C} & \dot{\mbf r} \\ \mbf 0^T & 0 \ebm
    = \bbm \mbs\omega^\times\mbf C & \dot{\mbf r} \\ \mbf 0^T & 0 \ebm
    = \bbm \mbs\omega^\times & \mbf v \\ \mbf 0^T & 0 \ebm\bbm \mbf C & \mbf r \\
    \mbf 0^T & 1 \ebm,
\end{equation}
where
\begin{equation}
  \mbf v = \dot{\mbf r} - \mbs\omega^\times\mbf r
\end{equation}
may be interpreted as the velocity of the reference point of a body as measured in the pretransformed frame but expressed in the body frame.  The last equality in (\ref{pose:1}) can be briefly written as
\begin{equation}\label{pose:3}
  \dot{\mbf T} = \mbs\upsilon^\wdg\mbf T,
\end{equation}
where
\begin{equation}
  \mbs\upsilon^\wdg = \bbm \mbs\omega^\times & \mbf v \\ \mbf 0^T & 0 \ebm
\end{equation}
for $\mbs\upsilon = [\,\mbf v^T\;\; \bfom^T\,]^T \in \Re^6$.  {\jobcolor (\citet{Muller:2021} has recently provided a review of the use of the exponential map, as well as the Cayley map, on $SE(3)$ for the integration of (\ref{pose:3}).)}  We make the same observation as above that for $\mbf T = \mbf T_{ba}$, $\mbs\upsilon = \mbs\upsilon_b^{ab}$ is the generalized velocity of $(O_a,\scF_a)$ relative to $(O_b,\scF_b)$ as referred to $(O_b,\scF_b)$.

As before, we may again express $\mbf T$ from (\ref{pose:3}) explicitly when $\mbs\upsilon$ is constant as
\begin{equation}
  \mbf T = \exp(\mbs\upsilon^\wdg t).
\end{equation}
Let us set $\bfxi = \mbs\upsilon t = \phi\mbs s$, where
\begin{equation}\label{mc:10}
  \mbs s  = \bbm p\mbf a + \mbf m^\times\mbf a \\ \mbf a \ebm
\end{equation}
is the constant {\em screw axis} of the Mozzi-Chasles theorem; $p$ is the (scalar) {\em pitch}, $\mbf a$ is again the Euler axis and $\mbf m$ is the {\em moment arm}.  (Note that $(\mbf m^\times\mbf a,\mbf a)$ are the {\em Pl\"ucker coordinates}.)  Then
\begin{equation}\label{pose:4}
  \mbf T(\bfxi)  = \exp\bfxi^\wdg
\end{equation}
is a general expression for $\mbf T \in SE(3)$ and $\bfxi^\wdg \in \se(3)$.

Finally, and following the same course, the kinematics for $\Tbig \in \Ad(SE(3))$ require
\begin{equation}\label{pose:5}
  \dot{\bsT} = \bbm \bfom^\times\bfC & (\dot{\bfr} + \bfr^\times\bfom^\times)\bfC \\
    \mbf O & \bfom^\times\bfC \ebm
    = \bbm \bfom^\times & \bfv^\times \\ \mbf O & \bfom^\times \ebm
    \bbm \bfC & \bfr^\times\bfC \\ \mbf O & \bfC \ebm.
\end{equation}
Again $\bfv = \dot{\bfr} - \bfom^\times\bfr$ and we have made use of the identity $(\bfom^\times\bfr)^\times = \bfom^\times\bfr^\times - \bfr^\times\bfom^\times$.   Hence
\begin{equation}\label{pose:6}
  \dot{\bsT} = \bfup^\Wdg\bsT,
\end{equation}
where we recall that
\begin{equation}
  \bfup^\Wdg = \bbm \bfom^\times & \bfv^\times \\ \mbf O & \bfom^\times \ebm.
\end{equation}
For this pose representation, we may write
\begin{equation}\label{pose:7}
  \Tbig(\bfxi) = \exp\bfxi^\Wdg,
\end{equation}
where $\bfxi^\Wdg$ belongs to $\mbox{ad}(\se(3))$.  Note that $\bfxi$ plays the same role for poses as $\mbs\phi$ does for rotations.

We may draw additional parallels among these three Lie groups resulting from the kinematics of these transformations.  The ``transport theorem'' for three-dimensional vectors, as is well known, may be expressed as
\begin{equation}
    \dot{\mbf r}_b = \bfC_{ba}\left(\dot{\mbf r}_a - {\mbs\om_a^{ba}}^\times\mbf r_a\right),
\end{equation}
which can be obtained directly by differentiating $\mbf r_b = \bfC_{ba}\mbf r_a$ while noting that $\mbs\om_b^{ab} = -\bfC_{ba}\mbs\om_a^{ba}$ and that Poisson's relation may be alternatively written as $\dot{\bfC}_{ba} = -\bfC_{ba}{\mbs\om_a^{ba}}^\times$ owing to the first of (\ref{pose:8}).  By the same route, the corresponding result for  homogeneous coordinates subject to transformations by elements of $SE(3)$ is
\begin{equation}
    \dot{\mbf s}_b = \mbf T_{ba}\left(\dot{\mbf s}_a - {\mbs\upsilon_a^{ba}}^\wdg\mbf s_a\right).
\end{equation}
When we turn to $\mbox{Ad}(SE(3))$ and generalized vectors, we must distinguish between those vectors that originate in $\mbox{ad}(\se(3))$ and those that originate in its dual $\mbox{ad}(\se^\ast(3))$.  In the former case, $\mbs t_b = \bsT_{ba}\mbs t_a$, which leads to
\begin{equation}
    \dot{\mbs t}_b = \bsT_{ba}\left(\dot{\mbs t}_a - {\mbs\upsilon_a^{ba}}^\Wdg\mbs t_a\right).
\end{equation}
However, in the latter case, $\mbs t_a = \bsT_{ba}^T\mbs t_b$ giving
\begin{equation}
    \dot{\mbs t}_a = \bsT_{ba}^T\left(\dot{\mbs t}_b + ({{\mbs\upsilon_b^{ab}}^\Wdg})^T\mbs t_b\right)
\end{equation}
as $\dot{\bsT}_{ba}^T = \bsT_{ba}^T({{\mbs\upsilon_b^{ab}}^\Wdg})^T$.

We mention that the exponential mapping in each of the cases above is only surjective as $\bfC, \mbf T$ and $\Tbig$ can be produced by many different values for $\bfph$ or $\bfxi$.  Other vector parameterizations of rotation and pose are possible as shown by \citet{bauchau03}, \citet{Bauchau:2011} and \citet{Barfoot&al:2021}.

It is not surprising that the nature of these representations, including the above identities, are intimately related to the eigenstructure, more generally a Jordan or block upper-triangular structure when degenerate eigenvalues are present, of the associated Lie algebras and Lie groups.

\input{so3}

\input{se3}

\input{adse3X}

\section{Concluding Remarks}

The eigenstructure of rotations and poses, analyzed through the Lie algebra of the corresponding Lie group, reveals much about their nature whether they are regarded as representations of the translational and rotational displacement of a body or as transformations.  Euler's classic results for rotations form the template for the analysis of pose transformations.  By the eigendecomposition of the latter we can likewise identify a {\em principal-axis pose}, that is, a translation along and rotation about a principal axis as in the Mozzi-Chasles theorem.  The repeated eigenvalues that arise in $\se(3)$ and $\ad(\se(3))$ appear to complicate matters; however, it is these eigenvalues, when unaccompanied by linearly independent eigenvectors, that account for the coupling of translation and rotation.

While the $SE(3)$ and $\Ad(SE(3))$ representations for pose have found broad use, the true sibling to $SO(3)$ for pose is $\Ad(SE(3))$.  Several commutator identities in $SO(3)$, as given in Table~\ref{table}, nicely carry over to $\Ad(SE(3))$.  Not so with $SE(3)$, where the identities sometimes need to be mediated by elements in $\Ad(SE(3))$.  {\jobcolor The Jacobian of $SO(3)$ extends naturally to $\Ad(SE(3)$ but there is no comparable Jacobian for $SE(3)$.}  The elements of $SO(3)$ and $\Ad(SE(3))$, moreover, respectively transform vectors in $\Re^3$ and $\Re^6$.  Those of $SE(3)$, despite representing both rotation and translation, do not transform vectors in $\Re^6$, only bound vectors in $\Re^3$.  Nevertheless, for pose representation, both $SE(3)$ and $\Ad(SE(3))$ have advantages as their wide application in many fields of science and engineering show.

Having taken this stroll through the eigenanalysis of rotations and poses, we hope to have shown the commonalities among the three Lie groups of their representations as well as identifying some aspects that are peculiar to each.

\input{arxivversion.bbl}

\end{document}

%% file: notation_header.tex
\newcommand{\diag}{\textup{diag}}
\newcommand{\bbm}{\begin{bmatrix}}
\newcommand{\ebm}{\end{bmatrix}}
\newcommand{\mbf}{\mathbf}
\newcommand{\mbs}[1]{{\boldsymbol{#1}}}

\def\om{\omega}

\newcommand{\beq}{\begin{equation}}
\newcommand{\eeq}{\end{equation}}
\newcommand{\bdis}{\begin{displaymath}}
\newcommand{\edis}{\end{displaymath}}

\newcommand{\bfr}{\mbf{r}}
\newcommand{\bfa}{\mbf{a}}

\newcommand{\bfC}{\mbf{C}}

\newcommand{\beqn}[1]{\begin{subequations}\label{eq:#1}\begin{eqnarray}}
\newcommand{\eeqn}{\end{eqnarray}\end{subequations}}
\newcommand{\wdg}{\wedge}

\newcommand{\Wdg}{\curlywedge}

\newcommand{\Tbig}{\;\mbox{\boldmath ${\cal T}$\unboldmath}}

\newcommand{\tr}{\mbox{tr}}

%% file: def.tex
\newcommand{\Ad}{\mbox{Ad}}
\newcommand{\ad}{\mbox{ad}}
\renewcommand{\Re}{\mathbb{R}}
\renewcommand{\so}{\mathfrak{so}}           
\newcommand{\se}{\mathfrak{se}}
\def \tr{\mbox{tr}\,}
\def \inv{{-1}}
\def \x{\times}

\newcommand{\0}{\mbf 0}
\newcommand{\1}{\mbf 1}

\newcommand{\bfD}{\mbf D}
\newcommand{\bfF}{\mbf F}

\newcommand{\bfJ}{\mbf J}
\newcommand{\bfO}{\mbf O}
\newcommand{\bfP}{\mbf P}
\newcommand{\bfT}{\mbf T}
\newcommand{\bfU}{\mbf U}
\newcommand{\bfV}{\mbf V}

\newcommand{\rp}{r}

\newcommand{\bfd}{\mbf d} 
\newcommand{\bfu}{\mbf u}
\newcommand{\bfv}{\mbf v}
\newcommand{\bfw}{\mbf w}
\newcommand{\bsd}{\mbi d}

\newcommand{\mbi}[1]{\textbf{\emph{#1}}}
\newcommand{\bsu}{\mbs{u}}

\newcommand{\bfPh}{\mbs\Phi}
\newcommand{\bfPs}{\mbs\Psi}
\newcommand{\bfUp}{\mbs\Upsilon}

\newcommand{\bsUp}{\bm{\mathit{\Upsilon}}}
\newcommand{\bsDe}{\bm{\mathit{\Delta}}}

\newcommand{\bfde}{\mbs\delta}
\newcommand{\bfet}{\mbs\eta}
\newcommand{\bfze}{\mbs\zeta}
\newcommand{\bfxi}{\mbs\xi}
\newcommand{\bfph}{\mbs\phi}
\newcommand{\bfrh}{\mbs\rho}
\newcommand{\bfup}{\mbs\upsilon}
\newcommand{\bfch}{\mbs\chi}
\newcommand{\bfom}{\mbs\omega}

\newcommand{\scF}{\mathcal F}
\newcommand{\bsD}{{\boldsymbol{\mathcal D}}}
\newcommand{\bsF}{{\boldsymbol{\mathcal F}}}
\newcommand{\bsJ}{{\boldsymbol{\mathcal J}}}
\newcommand{\bsT}{{\boldsymbol{\mathcal T}}}
\newcommand{\Dbig}{{\boldsymbol{\mathcal D}}}

\newcommand{\Ubig}{{\boldsymbol{\mathcal U}}}

\def\Tbig{\bsT}

\def\diag{\text{diag}}

\def\om{\omega}
\def\wdg{\wedge}
\def\Wdg{\times}

%% file: so3.tex
\section{Rotations---{\em SO}(3)}

The eigenstructure of rotation matrices has been well documented in the literature \citep{angeles02, markley14, bar00, ozdemir14}\ but we provide a summary of some key results that will lay the groundwork for the pose matrices.

For $\bfph^\times \in \so(3)$, the {\em characteristic equation}, or {\em eigenequation}, is
\begin{equation}
  \det(\lambda\1 - \bfph^\times) = \lambda(\lambda^2 + \phi^2) = 0.
\end{equation}
The eigenvalues are, in general, $\lambda = 0$, $i\phi, -i\phi$, where $i = \sqrt{-1}$.  The eigenvector corresponding to $\lambda = 0$ is of course $\mbf a$ ({\jobcolor because $\mbs\phi^\x\mbf a = \0 \equiv 0\mbf a$}), which is also an eigenvector of $\bfC$.  The case of $\phi = 0$ is trivial as it represents no rotation and $\bfC = \1$.

\subsection{Complex Decomposition}

For $\phi \neq 0$, $\bfph^\times$ is diagonalizable as the eigenvalues are all distinct.  The diagonalizing eigenmatrix can be verified to be
\begin{equation}
\bfU = \bbm \mbf{a} & \frac{1}{\sqrt{2}} \left(\mbf{b} - i \mbf{c}\right) & \frac{1}{\sqrt{2}} \left(\mbf{b} + i \mbf{c}\right) \ebm,
\end{equation}
where $\mbf{b}$ and $\mbf{c}$ are unit vectors that, with $\mbf a$, complete a dextral orthonormal set of basis vectors; in other words, $(\mbf a, \mbf b, \mbf c)$ satisfy  $\mbf{a}^\times \mbf{b} = \mbf{c}$, $\mbf{b}^\times\mbf{c} = \mbf{a}$, $\mbf{c}^\times\mbf{a} = \mbf{b}$.

Defining $\mbf D = \mbox{diag}[0,i\phi,-i\phi]$, the eigendecomposition of $\bfph^\times$ can be written as
\begin{equation}\label{so:1}
\mbs{\phi}^\times = \bfU \mbf{D} \bfU^H,
\end{equation}
where $\bfU^H$ denotes the Hermitian, i.e., the conjugate transpose, and is equal to $\bfU^{-1}$, i.e., $\mbf U$ is unitary.  It is noteworthy that $\mbf{D}$ depends only on the rotation angle, $\phi$, while $\bfU$ depends only on the rotation axis, $\mbf{a}$.

As $\bfU$ is orthogonal, it immediately follows that
\begin{equation}\label{so:1a1}
  \bfC = \exp\bfph^\times = \bfU(\exp\mbf D)\bfU^H
\end{equation}
and
\begin{equation}\label{so:1a2}
  \exp\mbf D = \diag\left[e^0,e^{i\phi},e^{-i\phi}\right]
  = \diag\bbm1, \cos\phi + i\sin\phi,
    \cos\phi - i\sin\phi\ebm
\end{equation}
using Euler's formula.  Hence
\begin{equation}
    \mbox{tr}\,\bfC = \mbox{tr}\exp\mbf D = 1 + 2\cos\phi, \qquad
    \det\bfC = \exp\mbox{tr}\,\mbf D = 1
\end{equation}
as is well known.

We can also recover the classic expression for the rotation matrix in terms of the Euler axis-angle variables by noting that, when multiplied out, $\bfC = \bfU(\exp\mbf D)\bfU^H$ yields
\begin{equation}
  \bfC = \mbf a\mbf a^T + \cos\phi(\mbf{b} \mbf{b}^T + \mbf{c} \mbf{c}^T)
    + \sin\phi(\mbf{c} \mbf{b}^T - \mbf{b} \mbf{c}^T).
\end{equation}
However, owing to the identities, $\mbf a\mbf a^T + \mbf{b} \mbf{b}^T + \mbf{c} \mbf{c}^T = \1$ and $\mbf{c} \mbf{b}^T - \mbf{b} \mbf{c}^T = (\mbf b^\times\mbf c)^\times = \mbf a^\times$, we arrive at
\begin{equation}\label{so:1a}
  \bfC = \cos\phi\,\1 + (1 - \cos\phi)\mbf a\mbf a^T + \sin\phi\,\mbf a^\times,
\end{equation}
which is the classic result.  We can also express this as
\begin{equation}
  \bfC = \1 + \sin\phi\,\mbf a^\times + (1 -\cos\phi)\mbf a^\times\mbf a^\times
\end{equation}
upon noting that $\mbf a\mbf a^T = \1 + \mbf a^\times\mbf a^\times$.

\subsection{Real Decomposition}

We can alternatively formulate the eigendecomposition in strictly real terms.  Any skew-symmetric matrix can be block-diagonalized by a real orthogonal eigenmatrix.  To be specific,
\begin{equation}\label{so:2}
  \bfph^\times = \bfUp\mbf d^\times\bfUp^T,
\end{equation}
where $\mbf d = [\,\phi \;\; 0 \;\; 0\,]^T = \phi\1_1$ ($\1_k$ is the $k$th column of the identity matrix) and
\begin{equation}
  \bfUp = \bbm \mbf a & \mbf b & \mbf c \ebm.
\end{equation}
Now
\begin{equation}\label{so:3}
  \bfC = \bfUp(\exp\mbf d^\times)\bfUp^T.
\end{equation}
Moreover,
\begin{equation}
  \exp\mbf d^\times = \bbm 1 & 0 & 0 \\ 0 & \cos\phi & -\sin\phi \\ 0 & \sin\phi & \cos\phi \ebm,
\end{equation}
which is the principal transformation matrix for a rotation about a principal ($x$) axis.  Expanding (\ref{so:3}) again produces (\ref{so:1a}).

We also observe that $\bfUp$ is an orthogonal matrix and because $(\mbf a,\mbf b,\mbf c)$ is dextral $\det\bfUp = 1$, which makes $\bfUp$ a rotation matrix, i.e., $\bfUp \in SO(3)$.  In fact, $\bfUp\mbf d = \bfph$.  From this observation, in conjunction with (\ref{so:3}), we have that $(\bfUp\mbf d)^\times = \bfUp\mbf d^\times\bfUp^T$ or equivalently $(\bfUp^T\bfph)^\times = \bfUp^T\bfph^\times\bfUp$, which are special instances of the first of (\ref{pose:8}).

The first identity of (\ref{pose:8}) may be shown in a variety of ways.  One route can make use of the latter eigendecomposition.  Consider $\bfC\bfph$, where both $\bfC$ and $\bfph$ are arbitrary.  The eigenvalues of $(\bfC\bfph)^\times$ are identical to those of $\bfph^\times$ so just as above we may write
\begin{equation}
  (\bfC\bfph)^\times = \bfPh\mbf d^\times\bfPh^T
\end{equation}
for some $\bfPh \in SO(3)$.  The eigenmatrix may be expressed as $\bfPh = \mbf R\bfUp$, where $\bfUp$ block-diagonalizes $\bfph^\times$ as in (\ref{so:2}) and as consequence $\mbf R \in  SO(3)$ as well.  Hence
\begin{equation}
  (\bfC\bfph)^\times = \mbf R\bfph^\times\mbf R^T.
\end{equation}
We now can argue that $\mbf R$ must be $\bfC$.  Postmultiplying by $\mbf R\bfph$ yields
\begin{equation}
  (\bfC\bfph)^\times\mbf R\bfph = \0.
\end{equation}
That is, $\mbf R\bfph$ must be parallel to $\bfC\bfph$; in other terms, $(\mbf R - \alpha\bfC)\bfph = \0$.  If $\bfph$ is arbitrary then $\mbf R = \alpha\bfC$ for some scalar $\alpha \in \Re$.  However, owing to $\mbf R$ being a rotation matrix, its determinant must be unity, which leads to the desired identity.

\begin{figure}
\centering
\includegraphics[width=0.9\textwidth]{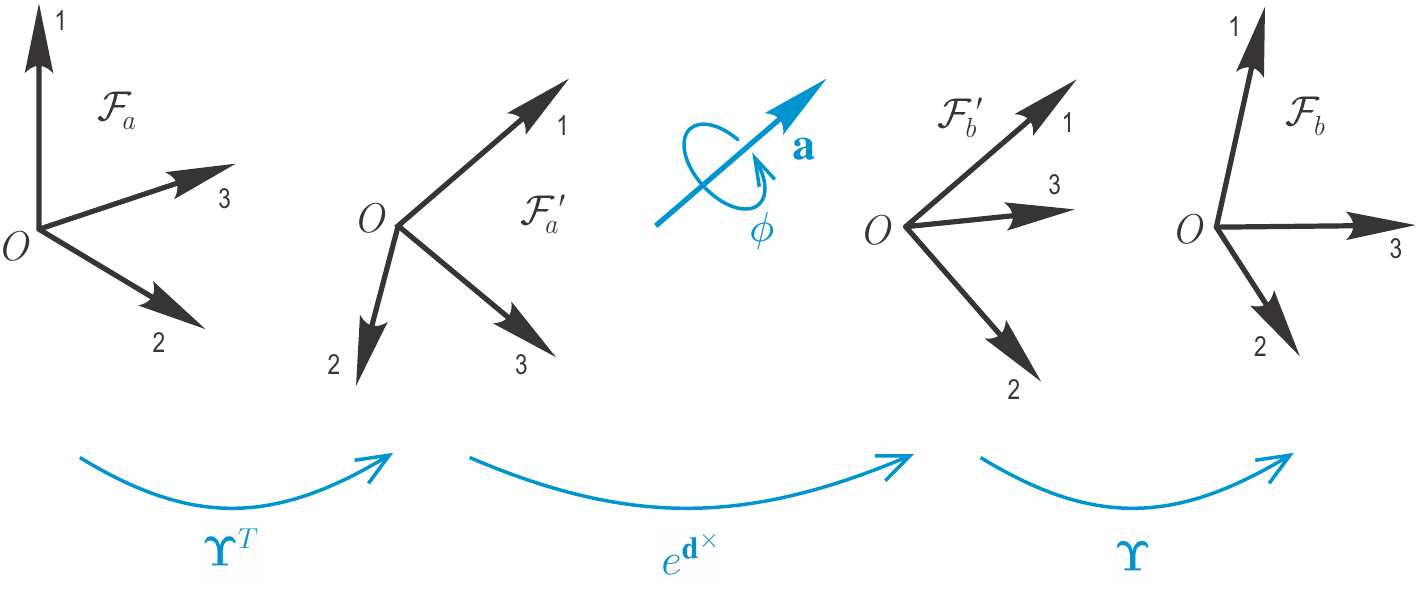}
\caption{A rotation in $SO(3)$ given in terms of its eigendecomposition.  The principal-axis rotation is $\phi$ (about $\mbf a$).}
\label{fig:so(3)}
\end{figure}

{\jobcolor
An arbitrary rotation decomposed in terms of its eigenparameters is illustrated in Figure~\ref{fig:so(3)}\ showing the sequence of three rotations as given in (\ref{so:3}).
}

\subsection{Minimal Polynomial and Characteristics of {\bf C}}

According to the Cayley-Hamilton theorem, every matrix satisfies its own characteristic equation and as such
\begin{equation}
  {\bfph^\times}^3 + \phi^2\bfph^\times = \mbf O,
\end{equation}
which is also the minimal polynomial for $\bfph^\times$ as all the eigenvalues (except for the trivial case of $\bfph = \0$) are distinct.  Note that the eigenvalues are given by the relation $\phi^2 = -\frac{1}{2}\tr{\mbs\phi^\times}^2$.

Because (\ref{so:1a1}) for $\bfC$ is a similarity transformation, the eigenvalues of $\bfC$ are those of $\exp\mbf D$ (or equivalently $\exp\mbf d^\times$ as (\ref{so:3}) is also a similarity transformation) and thus immediately recognized to be 1, $e^{i\phi}$, $e^{-i\phi}$.  The characteristic equation for $\bfC$ is accordingly
\begin{equation}
  (\lambda - 1)(\lambda - e^{i\phi})(\lambda - e^{-i\phi}) = \lambda^3 - \sigma\lambda^2 + \sigma\lambda - 1 = 0,
\end{equation}
where $\sigma = \mbox{tr}\,\bfC = 1 + 2\cos\phi$.  This can also be gleaned by the general formula for an (invertible) 3$\times$3 matrix:
\begin{equation}
    \lambda^3 - (\mbox{tr}\,\bfC)\lambda^2 + (\mbox{tr}\,\bfC^{-1})(\det\bfC)\lambda - (\det\bfC)\1 = \mbf O.
\end{equation}
We remark finally that the Cayley-Hamilton theorem yields
\begin{equation}\label{so:10}
   \mbf{C}^3 - (\mbox{tr}\,\mbf{C}) \, \mbf{C}^2 + (\mbox{tr}\,\mbf{C}) \, \mbf{C} - \mbf{1} = \mbf{O}
\end{equation}
as an interesting identity for $\bfC$ \citep{hughes86, markley14}.

{\jobcolor

\subsection{The Inverse of $\bfC$}

The inverse of the rotation matrix is given by its transpose.  In geometric terms, the inverse is obtained simply by reversing the Euler angle of rotation (keeping the Euler axis $\mbf a$ in the same direction), in short, by replacing $\phi$ with $-\phi$ or more generally $\mbs\phi$ with $-\mbs\phi$ so that
\begin{equation}\label{so3:10}
  \bfC^\inv = \bfC^T = \exp(-\phi\mbf a^\times) = \exp(-\mbs\phi^\times).
\end{equation}
This is reflected in the eigendecomposition as well because
\begin{equation}\label{so3:11}
   \bfC^\inv = \mbf U(\exp\mbf D)^\inv\mbf U^H = \mbf U\exp(-\mbf D)\mbf U^H
\end{equation}
and
\begin{equation}\label{so3:12}
    (\exp\mbf D)^\inv = \left(\text{diag}\left[1,e^{i\phi},e^{-i\phi}\right]\right)^\inv
    = \text{diag}\left[1,e^{-i\phi},e^{i\phi}\right] = \exp(-\mbf D).
\end{equation}
In the real decomposition,
\begin{equation}\label{so3:13}
    \bfC^\inv = \bfUp(\exp\mbf d^\times)^\inv\bfUp^T = \bfUp\exp(-\mbf d^\times)\bfUp^T.
\end{equation}
While this aspect of the inverse rotation matrix is well known, we shall see that the same interpretation applies to the pose transformation matrices of $SE(3)$ and $\Ad(SE(3))$.

}

\subsection{The Jacobian $\bfJ$}

An important quantity in rotational geometry and kinematics is the (left) {\it Jacobian} $\bfJ$ of $SO(3)$, which relates angular velocity to the Euler axis-angle variables: $\bfom = \bfJ\dot{\bfph}$.  The Jacobian is furthermore related to the rotation matrix by $\bfC = \1 + \bfph^\times\mbf J = \1 + \mbf J\bfph^\times$ and can be given explicitly as \citep{Pfister:1998,barfoot_ser17,condurache20}
\begin{equation}\label{eq:J}
  \mbf J = \sum_{k=0}^\infty \frac{{\bfph^\times}^k}{(k+1)!} = \1 + \frac{1 - \cos\phi}{\phi}\mbf a^\times + \frac{\phi - \sin\phi}{\phi}{\mbf a^\times}\mbf a^\times.
\end{equation}
But it can also be determined with the help of the foregoing eigendecomposition.  Note that the Jacobian is equally obtained by \citep{Park:1991,barfoot_ser17}
\begin{equation}\label{eq:Ja}
    \bfJ = \int_0^1\bfC^\alpha d\alpha.
\end{equation}
Using complex decomposition,
\begin{equation}
    \bfC^\alpha = \exp(\alpha\bfph^\x) = \bfU\exp(\alpha\mbf D)\bfU^H
\end{equation}
and
\begin{equation}\label{eq:intD}
    \int_0^1 \exp(\alpha\mbf D)d\alpha = \int_0^1 \text{diag}\left[1, e^{i\alpha\phi}, e^{-i\alpha\phi}\right]d\alpha
        = \text{diag}\left[\; 1,\; \frac{e^{i\phi} - 1}{i\phi},\; -\frac{e^{-i\phi} - 1}{i\phi} \;\right] = \mbf M.
\end{equation}
Thus
\begin{equation}
    \bfJ = \bfU\mbf M\bfU^H,
\end{equation}
which upon expansion can be shown to agree with (\ref{eq:J}).

We can do the same in real terms, which will be instructive when we consider the Jacobian in $\Ad(SE(3))$.  In this case,
\begin{equation}
    \bfC^\alpha = \bfUp\exp(\alpha\bfd^\x)\bfUp^T = \bfUp\exp(\alpha\phi\1_1^\x)\bfUp^T
\end{equation}
and
\begin{equation}\label{eq:intd}
    \int_0^1 \exp(\alpha\bfd^\x)d\alpha = \1_1\1_1^T - \frac{1}{\phi}\1_1^\x\left[\exp(\phi\1_1^\x) - \1\right]
    = \1 + \frac{1 - \cos\phi}{\phi}\1_1^\x + \frac{\phi - \sin\phi}{\phi}\1_1^\x\1_1^\x.
\end{equation}
We can again recover (\ref{eq:J}) by substituting into (\ref{eq:Ja}) and noting that $\bfUp\1_1^\x\bfUp^T = (\bfUp\1_1)^\x = \mbf a^\x$.

%% file: se3.tex
\section{Poses---{\em SE}(3)}

Let us now turn our attention to $SE(3)$.  As we will see, however, the analysis becomes more involved because we cannot always diagonalize elements of the Lie algebra or Lie group associated with poses and must therefore in general settle for an {\em upper-triangular decomposition} instead.

For an element of the Lie algebra for poses, $\bfxi^\wdg \in \se(3)$, the characteristic equation is
\begin{equation}\label{se:1}
\det \left(  \lambda \mbf{1} - \mbs{\xi}^\wdg \right) = \det \bbm \lambda \mbf{1} - \mbs{\phi}^\times & -\mbs{\rho} \\ \mbf{0}^T & \lambda  \ebm = \lambda^2 \left( \lambda^2 + \phi^2 \right) = 0.
\end{equation}
The four eigenvalues are $\lambda = 0, 0, i\phi, -i\phi$.  Three of these are shared by $\bfph^\times \in \so(3)$ but there is an extra copy of $\lambda = 0$.  The repeated eigenvalue may not allow for the diagonalization of $\bfxi^\wdg$.

There are three cases we can identify.  One is where there is only pure translation, no rotation, i.e., $\phi = 0, \bfrh \neq \0$ (the case where both are zero is trivial and will not be considered as $\mbf T = \1$).  In another, $\phi \neq 0$ but $\mbf a^T\mbs\rho = 0$.  This corresponds to both rotation and translation but the translation is restricted to be orthogonal to the axis of rotation; in other words, motion is confined to a plane.  In the final case, $\phi \neq 0$ and $\mbf a^T\mbs\rho \neq 0$, allowing for a general three-dimensional transformation.  While these three cases lead to different decompositions, the first two can be treated as special cases of the last, which is where we begin.

\subsection{Complex Decomposition}

The eigenvectors for the eigenvalues $\lambda = i\phi, -i\phi$ are simply those associated with the same eigenvalues for $\so(3)$ extended to $\Re^4$, i.e.,
\begin{equation}\label{se:01}
  \bfv_2 = \bbm \frac{1}{\sqrt{2}}(\mbf b - i\mbf c) \\ 0 \ebm, \qquad
    \bfv_3 = \bbm \frac{1}{\sqrt{2}}(\mbf b + i\mbf c) \\ 0 \ebm,
\end{equation}
where $\mbf b$ and $\mbf c$ are again orthogonal to $\bfph = \phi\mbf a$ with $(\mbf a,\mbf b,\mbf c)$ forming a dextral orthonormal basis.  (We keep the same ordering as in the rotational case.)

As for $\lambda = 0$, we seek $\bfv = [\, \bfu^T \;\; w\,]^T$ such that
\begin{equation}
  \bfxi^\wdg\bfv = \bbm \bfph^\times & \bfrh \\ \0^T & 0 \ebm \bbm \bfu \\ w \ebm
    = \bbm \bfph^\times\bfu + w\bfrh \\ 0 \ebm = \0.
\end{equation}
We can easily verify that
\begin{equation}\label{se:5}
  \bfv = \bbm \mbf a \\ 0 \ebm
\end{equation}
is an eigenvector.  But, unfortunately, $\bfxi^\wdg$ does not surrender another (linearly independent) eigenvector (unless $\mbf a^T\bfrh = 0$ with $\phi \neq 0$).  We can see this immediately because the fourth eigenvector would have to have $w \neq 0$ for linear independence.  This requires $\bfu$ such that $\bfph^\times\bfu$ is nonzero to eliminate the $w\bfrh$ term but $\bfph^\times\bfu$ would then have to be orthogonal to $\bfph$, i.e., to $\mbf a$, and in general $\bfrh$ is not in this direction.

We accordingly look for a generator $\bfv_4$ such that
\begin{equation}
  {\bfxi^\wdg}^2\bfv_4 = \bbm \mbs{\phi}^\times\mbs{\phi}^\times & \mbs{\phi}^\times\mbs{\rho} \\ \mbf{0}^T & 0 \ebm \bbm \bfu_4 \\ w_4 \ebm
  = \bbm \mbs{\phi}^\times\mbs{\phi}^\times \bfu_4 + w_4 \mbs{\phi}^\times\mbs{\rho} \\ 0 \ebm = \mbf{0},
\end{equation}
which is satisfied by
\begin{equation}
  \bfv_4 = \bbm \phi^{-1}\mbf a^\times\bfrh \\ 1 \ebm.
\end{equation}
A second linearly independent eigenvector is $\bfv_1$ such that
\begin{equation}
  \rp\bfv_1 = \bfxi^\wdg\bfv_4,
\end{equation}
which in fact takes us back to (\ref{se:5}) with $\rp = \mbf a^T\bfrh$.  This is of course an application of the Jordan procedure generating a chain of two vectors $\{\bfv_1,\bfv_4\}$ \citep{Gantmacher:1959}.  However, instead of producing the canonical Jordan normal form we obtain an upper-triangular decomposition; replacing Jordan's off-diagonal entry of unity with $\rp$ allows us to treat the more specific cases as special instances of the more general one.

It may be verified that
\begin{equation}
  \bfxi^\wdg\bfV = \bfV\mbf E,
\end{equation}
where
\begin{equation}\label{se:10}
\begin{gathered}
  \mbf E = \bbm \mbf D & \rp\1_1 \\ \0^T & 0 \ebm,\\
  \bfV = \bbm \mbf U & \bfu_4 \\ \0^T & 1 \ebm = \bbm \mbf a & \mbf e & \bar{\mbf e} & \phi^{-1}\mbf a^\times\bfrh \\ 0 & 0 & 0 & 1 \ebm
\end{gathered}
\end{equation}
and for brevity $\mbf e = \frac{1}{\sqrt{2}}(\mbf b - i\mbf c)$ while $\mbf D$ is as in the rotational case; $\bar{\mbf e}$ indicates the complex conjugate.

The decomposition of $\bfxi^\wdg$ is then given by
\begin{equation}
  \bfxi^\wdg = \bfV\mbf E\bfV^{-1}
\end{equation}
and the transformation matrix $\mbf T$ by
\begin{equation}\label{se:4}
  \mbf T = \bfV(\exp\mbf E)\bfV^{-1} = \bbm \bfC & (\1 - \bfC)\bfu_4 + \rp\mbf a \\
        \0^T & 1 \ebm
\end{equation}
given that
\begin{equation}
  \exp\mbf E = \bbm \exp\mbf D & \rp\1_1 \\ \0^T & 1 \ebm,
\end{equation}
and $\bfC = \mbf U(\exp\mbf D)\mbf U^H$.  However, recalling that $\bfC = \1 + \bfJ\mbs\phi^\x = \1 + \phi\bfJ\mbf a^\x$, the upper-right block in (\ref{se:4}) can be simplified by
\begin{equation}
\left( \mbf{1} - \mbf{C} \right) \bfu_4 = \left( -\phi \mbf{J} \mbf{a}^\times \right) \frac{1}{\phi} \mbf{a}^\times \mbs{\rho}
= \mbf{J} \left( \mbf{1} - \mbf{a} \mbf{a}^T \right) \mbs{\rho} = \mbf{J} \mbs{\rho} - \rp\mbf a.
\end{equation}
Substituting back, we have
\begin{equation}\label{se:4b}
\mbf{T} = \bbm \mbf{C} & \mbf{J} \mbs{\rho} \\ \mbf{0}^T & 1 \ebm,
\end{equation}
the standard form.

\paragraph{Special Case I: $\mbs\phi = \0$, $\mbs\rho \neq \0$.}
As mentioned above, we can treat the situation of a pure translation or a rotation and translation confined to a plane as special cases of the foregoing.  For $\phi = 0$, corresponding to a pure translation,
\begin{equation}
  \bfxi^\wdg = \bbm \mbf O & \bfrh \\ \0^T & 0 \ebm,
\end{equation}
whose eigenvalues are $\lambda = 0$ repeated four times.  However, $\mbox{rank}\,\bfxi^\wdg = 1$, indicating that $\bfxi^\wdg$ cannot be diagonalized and that Jordan decomposition is in order.  Given that $\mbox{rank}\,{\bfxi^\wdg}^k = 0$ for $k \geq 2$, we require a generalized eigenvector of rank two just as in the foregoing.  The general framework can indeed accommodate this special case if we select the ``rotation'' axis, which can be arbitrary as there is no rotation, to be $\rho^{-1}\bfrh$ with $\phi \rightarrow 0$.  The upper-triangular matrix $\mbf E$ and the eigenmatrix $\bfV$, by (\ref{se:10}), become
\begin{equation}
  \mbf E = \bbm \bfO & \rho\1_1 \\ \0^T & 0 \ebm, \qquad
  \bfV = \bbm \bfU & \0 \\ \0^T & 1 \ebm =  \bbm \mbf a & \mbf e & \bar{\mbf e} & \0 \\ 0 & 0 & 0 & 1 \ebm
\end{equation}
as may be verified by performing the eigenanalysis {\em ab initio}.  Note that in this case it is actually not necessary to introduce complex eigenvectors. (As a result, we may replace $\mbf e$ and $\bar{\mbf e}$ with $\mbf b$ and $\mbf c$.)  Moreover, we retrieve
\begin{equation}
  \mbf T = \bbm \1 & \bfrh \\ \0^T & 1 \ebm
\end{equation}
from (\ref{se:4}), as expected.

\paragraph{Special Case II: $\mbs\phi \neq \0$, $\mbf a^T\mbs\rho = 0$.}
For the case of a rotation and translation confined to a plane, $\rp = \mbf a^T\bfrh = 0$, indicating that $\bfxi^\wdg$ is diagonalizable and indeed it is because $\mbox{rank}\,\bfxi^\wdg = 2$, allowing for two linearly independent eigenvectors.  But again the general framework above handles this case.  Equation (\ref{se:10}) still gives the eigenmatrix $\bfV$ and $\mbf E$ is diagonal, reducing to $\mbf E = \mbox{diag}\,[0, i\phi, -i\phi, 0]$.  The transformation $\mbf T$ remains (\ref{se:4b}).

\subsection{Real Decomposition}

As we did with $SO(3)$, we can analyze the eigenstructure of $SE(3)$ in purely real terms.  The real eigenvectors can again be immediately inferred from the complex eigenvectors of (\ref{se:01}), from which we may write
\begin{equation}\label{se:4a}
  \bfxi^\wdg = \bfPs\bfde^\wdg\bfPs^{-1},
\end{equation}
where
\begin{equation}
\begin{gathered}
    \bfde = \bbm r\1_1 \\ \bfd \ebm = \bbm r\1_1 \\ \phi\1_1 \ebm, \qquad\bfde^\wdg = \bbm \mbf d^\times & \rp\1_1 \\ \0^T & 0 \ebm
    = \bbm 0 & 0 & 0 & r \\
    0 & 0 & -\phi & 0 \\ 0 & \phi & 0 & 0 \\ 0 & 0 & 0 & 0 \ebm,\\
  \bfPs = \bbm \bfUp & \bfu_4 \\ \0^T & 1 \ebm
    = \bbm \mbf a & \mbf b & \mbf c & \phi^{-1}\mbf a^\times\bfrh \\ 0 & 0 & 0 & 1 \ebm.
\end{gathered}
\end{equation}
Again, $\mbf d$ and $\bfU$ are recycled from the rotational case.

The transformation $\mbf T$ is still of the form,
\begin{equation}\label{se:11}
  \mbf T  = \bfPs(\exp\bfde^\wdg)\bfPs^{-1},
\end{equation}
where
\begin{equation}\label{se:12}
  \exp\bfde^\wdg = \bbm \mbf \mbf \exp\mbf d^\times & \rp\1_1 \\ \0^T & 1 \ebm
    = \bbm 1 & 0 & 0 & \rp \\ 0 & \cos\phi & -\sin\phi & 0 \\
    0 & \sin\phi & \cos\phi & 0 \\ 0 & 0 & 0 & 1 \ebm,
\end{equation}
which is itself an element of $SE(3)$ while $\bfde^\wdg$ belongs to $\se(3)$; in fact, $\exp\bfde^\wdg$ represents a principal-axis rotation and translation about and along the same ($x$) axis, which we may refer to as a {\it principal-axis pose}.  We observe that $\bfPs$ is also an element of $SE(3)$, with a translation perpendicular to $\mbf a$ and $\bfrh$.

This also reduces to the special cases of a pure translation and a planar pose transformation (rotation and translation confined to a plane).

\begin{figure}
\centering
\includegraphics[width=0.94\textwidth]{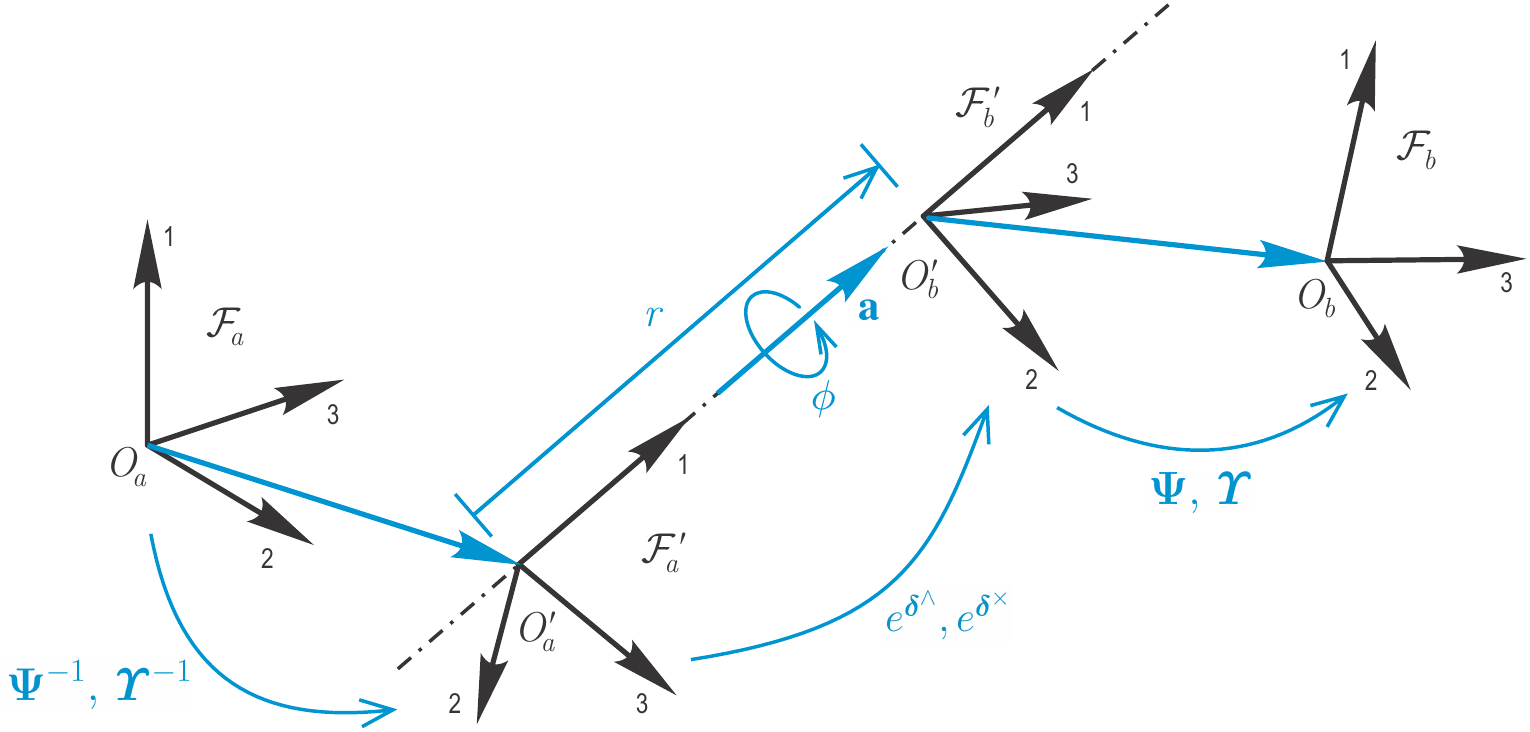}
\caption{A pose transformation given in terms of its eigendecomposition using $\bfPs$ and $e^{\bfde^\wdg}$ in $SE(3)$ and $\bsUp$ and $e^{\bfde^\x}$ in $\text{\normalfont Ad}(SE(3))$.  The principal-axis pose is given by the rotation $\phi$ (about $\mbf a$) and the translation $r$ (along $\mbf a$).}
\label{fig:se(3)}
\end{figure}

{\jobcolor
An arbitrary pose transformation in $SE(3)$ is depicted in Figure~\ref{fig:se(3)}\ in terms of its eigendecomposition.
}

\subsection{Minimal Polynomial and Characteristics of $\mbf T$}

The minimal polynomial for elements $\bfxi^\wdg$ in $\se(3)$, given in general the lack of a complete set of linearly independent eigenvectors for the repeated eigenvalue $\lambda = 0$, is
\begin{equation}
  {\bfxi^\wdg}^4 + \phi^2{\bfxi^\wdg}^2 = \mbf O.
\end{equation}
However, in the case of a pure translation it simplifies to ${\bfxi^\wdg}^2 = \mbf O$ and in the case of a planar pose to ${\bfxi^\wdg}^3 + \phi^2{\bfxi^\wdg} = \mbf O$.  We add that in general $\phi^2 = -\frac{1}{2}\tr{\mbs\phi^\times}^2 = -\frac{1}{2}\tr{\bfxi^\wdg}^2$ \citep{selig07}.

The eigenstructure of $\se(3)$ of course is reflected in the eigenstructure of $SE(3)$.  Given that (\ref{se:4}) is a similarity transformation, the eigenvalues of $\mbf T$ are the exponentials of the eigenvalues of $\bfxi^\wdg$, i.e., the eigenvalues are $1, 1, e^{i\phi}, e^{-i\phi}$.  Moreover,
\begin{equation}
  \mbf T\bfV = \bfV\exp\mbf E
\end{equation}
and, as $\exp\mbf E$ is an upper-triangular decomposition, $\bfV$ is the complex eigenmatrix for $\mbf T$.  Likewise, $\bfPs$ is the real eigenmatrix because the central $2\times 2$ matrix in (\ref{se:12}) accounts for the eigenvalues $e^{i\phi}$ and $e^{-i\phi}$.

The characteristic equation for $\mbf T$ is given by
\begin{equation}
  \det(\lambda\1 - \mbf T) = (\lambda - 1)\det(\lambda\1 - \mbf C) = 0,
\end{equation}
which becomes
\begin{equation}
  \lambda^4 - (1 + \sigma)\lambda^3 + 2\sigma\lambda^2 - (1 + \sigma)\lambda + 1 = 0
\end{equation}
recalling $\sigma = \tr\mbf C = 1 + 2\cos\phi$.  As $\tr\mbf T = 1 + \tr\mbf C = 2(1 + \cos\phi)$,  the Cayley-Hamilton theorem reveals that
\begin{equation}\label{se:20}
  \mbf T^4 - (\tr\mbf T)\mbf T^3 - 2(1 - \tr\mbf T)\mbf T^2 - (\tr\mbf T)\mbf T + \1 = \mbf O,
\end{equation}
which may be compared to (\ref{so:10}).

{\jobcolor

\subsection{The Inverse of $\mbf T$}

Given the exponential map for $\mbf T$, we have immediately that its inverse is
\begin{equation}\label{se3:10}
    \mbf T^\inv = (\exp\mbs\xi^\wdg)^\inv = \exp(-\mbs\xi^\wdg).
\end{equation}
Using complex decomposition, we have
\begin{equation}\label{se3:11}
    \mbf T^\inv = \mbf V(\exp\mbf E)^\inv\mbf V^\inv = \mbf V\exp(-\mbf E)\mbf V^\inv,
\end{equation}
where
\begin{equation}\label{se3:12}
    (\exp\mbf E)^\inv = \exp\bbm -\mbf D & -r\1_1 \\ \0^T & 0 \ebm \equiv \exp(-\mbf E).
\end{equation}
Noting again that $\mbf D = \text{diag}\,[0,i\phi,-i\phi]$, we see that the inverse pose transformation is given by reversing $\phi$ and $r$ as may be expected.  This fact may also be gleaned from the real decomposition as
\begin{equation}\label{se3:13}
    \mbf T^\inv = \bfPs(\exp\bfde^\wdg)^\inv\bfPs^\inv = \bfPs\exp(-\bfde^\wdg)\bfPs^\inv
\end{equation}
and
\begin{equation}\label{se3:14}
    (\exp\bfde^\wdg)^\inv = \exp\bbm -\mbf d^\x & -r\1_1 \\ \0^T & 0 \ebm \equiv \exp(-\bfde^\wdg).
\end{equation}
As we can see, the construction of the inverse pose transformation recalls the $SO(3)$ case but instead of flipping just the sign of $\phi$ one must flip the sign of both $\phi$ and $r$.

}

%% file: adse3X.tex
\section{Pose Adjoints---\mbox{\large\normalfont Ad}({\em SE}(3))}

We now address pose adjoints, the 6$\times$6 representation of $\Ad(SE(3))$.  For $\bfxi^\Wdg \in \ad(\se(3))$, the characteristic equation is
\begin{equation}
  \det \left(  \lambda \mbf{1} - \mbs{\xi}^\Wdg \right) = \det \bbm \lambda \mbf{1} - \mbs{\phi}^\times & -\mbs{\rho}^\times \\ \mbf O & \lambda \mbf{1} - \mbs{\phi}^\times  \ebm  = \left( \det \left(  \lambda \mbf{1} - \mbs{\phi}^\times \right) \right)^2 = \lambda^2\left( \lambda^2 + \phi^2\right)^2  = 0.
\end{equation}
The eigenvalues in general are $\lambda = 0, i\phi, -i\phi$, each of which is repeated twice, essentially two copies of those in the rotational case.  As may be anticipated, the analysis parallels that of $SE(3)$.  Once again, the diagonalizability of $\bfxi^\Wdg$ requires $\mbf a^T\bfrh = 0$.  But this, the case of rotation and translation confined to a plane, and the case of pure translation can be treated as special cases of the general problem, which we now address.

\subsection{Complex Decomposition}

The eigenvalues $\lambda = 0$ do come with linearly independent eigenvectors.  These may be taken as
\begin{equation}\label{eig:1}
  \bsu_1 = \bbm \mbf a \\ \0 \ebm, \qquad
  \bsu_4 = \bbm \phi^{-1}\bfrh \\ \mbf a \ebm.
\end{equation}
By a similar argument to that used for $\se(3)$, we can conclude that neither $\lambda = i\phi$ nor $\lambda = -i\phi$ has two linearly independent eigenvectors.  We therefore search for generators in the kernel of $(\pm i\phi\1 - \bfxi^\Wdg)^2$.

For $\lambda = i\phi$, the eigenvectors are spanned by
\begin{equation}\label{eig:1a}
  \bsu =  \bbm \mbf e \\ \mbf{0} \ebm,
\end{equation}
where $\mbf e = \frac{1}{\sqrt{2}}(\mbf b - i\mbf c)$.  A linearly independent generator to this vector is
\begin{equation}\label{eig:2}
\bsu_5 = \bbm i\phi^{-1} \mbf e^\times \mbs{\rho} \\  \mbf e \ebm.
\end{equation}
The companion eigenvector is then
\begin{equation}
  -i\rp\bsu_2 = (i\phi\1 - \bfxi^\Wdg)\bsu_5 = -i\rp\bbm \mbf e \\ \0 \ebm,
\end{equation}
where again $\rp = \mbf a^T\bfrh$.  Thus $\bsu_2$ is indeed given by (\ref{eig:1a}).  For $\lambda = -i\phi$, the generator is $\bsu_6 = \bar{\bsu}_5$ and its companion is $\bsu_3 = \bar{\bsu}_2$, where $i\rp\bsu_3 = (-i\phi\1 - \bfxi^\Wdg)\bsu_6$.

Choosing the above scaling of eigenvectors allows for the decomposition of $\bfxi^\Wdg$ as
\begin{equation}\label{eig:3}
  \bfxi^\Wdg = \Ubig\Dbig\Ubig^{-1},
\end{equation}
where
\begin{equation}\label{eig:4}
  \Dbig = \bbm \mbf D & \bfF \\ \mbf O & \mbf D \ebm, \qquad
  \Ubig = \bbm \mbf U & \mbf W \\ \mbf O & \mbf U \ebm
\end{equation}
and
\begin{equation}\label{eig:5}
  \bfF = \bbm 0 & 0 & 0 \\ 0 & i\rp & 0 \\ 0 & 0 & -i\rp \ebm, \quad
  \mbf W = \phi^{-1}\bbm \bfrh & i\mbf e^\times\bfrh & -i\bar{\mbf e}^\times\bfrh \ebm
\end{equation}
with $\mbf D$ and $\mbf U$ as in the rotational case.

To recover the transformation $\bsT$, we proceed as before, namely,
\begin{equation}\label{eig:5}
  \Tbig = \exp\bfxi^\Wdg = \Ubig(\exp\Dbig)\Ubig^{-1} = \bbm \mbf C & \mbf X \\
    \mbf O & \mbf C \ebm,
\end{equation}
where
\begin{equation}
    \exp\Dbig = \bbm \exp\mbf D & (\exp\mbf D)\mbf F \\ \bfO & \exp\mbf D \ebm
\end{equation}
and
\begin{equation}
  \mbf X = \mbf W\mbf U^H\mbf C - \mbf C\mbf W\mbf U^H + \mbf U\bfF\mbf U^H\mbf C
\end{equation}
upon noting that $(\exp\mbf D)\bfF = \bfF\exp\mbf D$.  It can be shown with a little effort that this last quantity reduces to
\begin{equation}
  \mbf X = (\mbf J\bfrh)^\times\mbf C.
\end{equation}
It will be helpful to recall that $\mbf C = \1 + \phi\mbf a^\times\mbf J$ and to observe the identities $\mbf e^\times\mbf a = -i\mbf e$, $\mbf e^T\mbf a = 0$ and $\mbf e^\times\mbf a^\times = -i\mbf e^\times + \mbf e\mbf a^T$ given that $(\mbf a, \mbf b, \mbf c)$ is a dextral orthonormal set of vectors.  As such, (\ref{eig:5}) becomes
\begin{equation}\label{eig:20}
  \Tbig = \bbm \mbf C & (\mbf J\bfrh)^\times\mbf C \\ \mbf O & \mbf C \ebm,
\end{equation}
as would be anticipated.

We make the observation that, as the eigenvectors of $\bfxi^\times$ are shared with $\bsT$, the eigenvectors $\mbs u_1$ and $\mbs u_4$ given in (\ref{eig:1}) correspond to the unit eigenvalue.  Any linear combination of the two is a screw axis of Mozzi and Chasles.  In fact,
\begin{equation}
  \beta\mbs u_1 + \mbs u_4 = \bbm (\beta + \phi^{-1}\mbf a^T\mbs\rho)\mbf a + \phi^{-1}(\mbf a^\times\mbs\rho)^\times\mbf a \\ \mbf a \ebm,
\end{equation}
where, in comparing with (\ref{mc:10}), we may identify $p = \beta + \phi^{-1}\mbf a^T\mbs\rho$ and $\mbf m = \phi^{-1}\mbf a^\times\mbs\rho$.  For an exhaustive treatment of the Mozzi-Chasles theorem in the context of the $\mbox{Ad}(SE(3))$ representation, consult \citet{Bauchau:2011}.

\paragraph{Special Cases.}
The case of a rotation and translation confined to a plane is subsumed by the foregoing development.  The eigenvectors are the same but $\bfxi^\times$ is diagonalizable; the off-diagonal entries in $\Dbig$ vanish as $\rp = \mbf a^T\bfrh = 0$.

For a pure translation,
\begin{equation}
  \bfxi^\Wdg = \bbm \mbf O & \bfrh^\times \\ \mbf O & \mbf O \ebm,
\end{equation}
all the eigenvalues are zero but the rank of $\bfxi^\Wdg$ is 2 provided $\bfrh \neq \0$.  In this case, as in $\se(3)$, we can set $\mbf a$ to be aligned with $\bfrh$ such that $\phi^{-1}\bfrh \rightarrow \mbf a$ and as a result $r = \rho$, the magnitude of $\bfrh$.

\subsection{Real Decomposition}

Once again we can recast the foregoing complex analysis completely in real terms.  The decomposition of $\bfxi^\Wdg$ as can be easily shown is
\begin{equation}\label{adre:1}
  \bfxi^\Wdg = \bsUp_0\bfde^\Wdg\bsUp_0^{-1},
\end{equation}
where $\bfde$ is defined as before,
\begin{equation}\label{adse:1a}
  \bfde = \bbm r\1_1 \\ \mbf d \ebm = \bbm r\1_1 \\ \phi\1_1 \ebm, \qquad
  \bsUp_0 = \bbm \bfUp & \bfP_0 \\ \mbf O & \bfUp \ebm,
\end{equation}
$\mbf d$ and $\bfUp$ are recycled from the foregoing $SO(3)$ and $SE(3)$ developments and \begin{equation}\label{adre:2}
  \bfP_0 = \phi^{-1}\bbm\bfrh & \mbf c^\times\bfrh & -\mbf b^\times\bfrh\ebm.
\end{equation}
That is, the generalized eigenvectors are
\begin{equation}
\begin{gathered}
  \bfup_1 = \bbm \mbf a \\ \0 \ebm = \bsu_1, \qquad
  \bfup_2 = \bbm \mbf b \\ \0 \ebm, \qquad
  \bfup_3 = \bbm \mbf c \\ \0 \ebm, \\
  \bfup_4 =\bbm \phi^{-1}\bfrh \\ \mbf a \ebm = \bsu_4, \qquad
  \bfup_5 = \bbm \phi^{-1}\mbf c^\times\bfrh \\ \mbf b \ebm, \qquad
  \bfup_6 = \bbm -\phi^{-1}\mbf b^\times\bfrh \\ \mbf c \ebm.
\end{gathered}
\end{equation}
These eigenvectors are obtained by parsing the complex eigenvectors above; however, these aren't our preferred choice.  We observe that $\bfP_0$ can be written as
\begin{equation}\label{eig:001}
  \bfP_0 = \phi^{-1}\left[\rp\1 + (\mbf a^\times\bfrh)^\times\right]\bfUp.
\end{equation}
This unfortunately does not admit $\bsUp_0$ in $\Ad(SE(3))$.  The $\rp\1$ term prevents it  but this term can be dropped without consequence as we shall see.

The transformation matrix based on (\ref{adre:1}) is
\begin{equation}\label{adre:4}
  \Tbig = \bsUp_0(\exp\bfde^\Wdg)\bsUp_0^{-1},
\end{equation}
where
\begin{equation}
  \exp\bfde^\Wdg = \bbm \mbf C_1 & \mbf C_1(\rp\1_1)^\times \\
    \mbf O & \mbf C_1 \ebm
    = \bbm \mbf C_1 & \rp\1_1^\times\mbf C_1 \\
    \mbf O & \mbf C_1 \ebm
\end{equation}
with $\mbf C_1 = \exp\mbf d^\times = \exp(\phi\1_1^\times)$, which is a principal ($x$ axis) rotation, and we have noted that $\mbf C_1\1_1^\times = (\mbf C_1\1_1)^\times\mbf C_1 = \1_1^\times\mbf C_1$.  We also recognize $\exp\bfde^\Wdg$ as a principal-axis pose, i.e., a principal rotation about and a principal translation along the same ($x$) axis through the angle $\phi$ and the distance $\rp$.

For completeness, we can recover (\ref{eig:20}) using this formulation as well.  The upper-right block of (\ref{adre:4}) is
\begin{equation}\label{eig:002}
  \mbf X = \phi^{-1}(\mbf a^\times\bfrh)^\times\mbf C - \phi^{-1}\mbf C(\mbf a^\times\bfrh)^\times + \rp\mbf a^\times\mbf C,
\end{equation}
where $\mbf C = \bfUp\mbf C_1\bfUp^T = \bfUp(\exp\mbf d^\times)\bfUp^T$.  The first two terms reduce to
\begin{equation}
  \phi^{-1}(\mbf a^\times\bfrh)^\times\mbf C - \phi^{-1}\mbf C(\mbf a^\times\bfrh)^\times
   = \phi^{-1}[(\1 - \mbf C)\mbf a^\times\bfrh]^\times\mbf C
   = -(\mbf J\mbf a^\times\mbf a^\times\bfrh)^\times\mbf C
   = (\mbf J\bfrh)^\times\mbf C - \rp\mbf a^\times\mbf C,
\end{equation}
from which (\ref{eig:20}) follows.

The disrupting $r\1$ term in (\ref{eig:001}) leaves no residual trace in (\ref{eig:002}).  We therefore should expect that replacing $\bfP_0$ with
\begin{equation}\label{eig:003}
  \bfP = \phi^{-1}(\bfa^\times\bfrh)^\times\bfUp
\end{equation}
leads to
\begin{equation}\label{eig:004}
  \Tbig = \bsUp(\exp\bfde^\Wdg)\bsUp^{-1}
\end{equation}
where
\begin{equation}
   \bsUp = \bbm \bfUp & \bfP \\ \bfO & \bfUp \ebm = \bbm \bfUp & \phi^\inv(\bfa^\times\bfrh)^\times\bfUp \\ \bfO & \bfUp \ebm.
\end{equation}
And indeed it can be straightforwardly verified that $\bfxi^\Wdg\!\bsUp = \bsUp\bfde^\Wdg$, demonstrating that $\bsUp$ is a proper eigenmatrix for $\bfxi^\Wdg$.  This can be explained by essentially taking a different linear combination of the eigenvectors for each pair of eigenvalues.

The decomposition given by (\ref{eig:004}) consists entirely of elements in $\Ad(SE(3))$ as now $\bsUp$ is also a pose-adjoint transformation.

{\jobcolor
We may again refer to Figure~\ref{fig:se(3)}\ for a geometric illustration of an arbitrary pose transformation in $\Ad(SE(3))$ in terms of its eigendecomposition.
}

\vspace*{10pt}

\stepcounter{subsection}
\noindent
{\bf \arabic{section}.\arabic{subsection}\,\, Minimal Polynomial and Characteristics of $\bsT$}%
\vspace*{8pt}

\noindent
The repeated eigenvalue $\lambda = 0$ always comes with two linearly independent eigenvectors; however, in general, $\lambda = \pm i\phi$ do not.  In such cases, the minimal polynomial for elements $\bfxi^\Wdg$ in $\ad(\se(3)$ is \citep{Borri&al:2000}
\begin{equation}
  \bfxi^\Wdg({\bfxi^\Wdg}^2 + \phi^2\1)^2 = \mbf O.
\end{equation}
In the case of pure translation, in which all the eigenvalues are zero, the minimal polynomial is ${\bfxi^\Wdg}^2 = \mbf O$ and, for the case of a planar pose, in which a complete set of eigenvectors exists, the minimal polynomial is $\bfxi^\Wdg({\bfxi^\Wdg}^2 + \phi^2\1) = \mbf O$.  We also note that $\phi^2 = -\frac{1}{2}\tr{\mbs\phi^\times}^2 = -\frac{1}{4}\tr{\bfxi^\Wdg}^2$.

It is interesting that the troublesome eigenvalue in $\se(3)$ is $\lambda = 0$ whereas in $\ad(\se(3))$ these eigenvalues are not degenerate but $\lambda = \pm i\phi$ are.  Nevertheless, in both algebras the degeneracy is caused by $\mbf a^T\bfrh \neq 0$.

For completeness, we mention that
\begin{equation}
  \Tbig^5 + (1 - \mbox{tr}\Tbig)\Tbig^4 + (1 + \mbox{$\frac{1}{4}$}\mbox{tr}^2\!\Tbig)\Tbig^3
    - (1 + \mbox{$\frac{1}{4}$}\mbox{tr}^2\!\Tbig)\Tbig^2 - (1 - \mbox{tr}\Tbig)\Tbig - \1 = \bfO
\end{equation}
which may be compared to (\ref{so:10}) for $\bfC$ and (\ref{se:20}) for $\mbf T$.


\vspace*{10pt}

{\jobcolor

\stepcounter{subsection}
\noindent
{\bf \arabic{section}.\arabic{subsection}\,\, The Inverse of $\bsT$}%
\vspace*{8pt}

\noindent
We may follow in the footsteps of the $SO(3)$ and $SE(3)$ developments to find that the inverse of $\bsT$ is
\begin{equation}\label{adse3:10}
    \bsT^\inv = (\exp\bfxi^\times)^\inv = \exp(-\bfxi^\times).
\end{equation}
Again, using complex decomposition,
\begin{equation}\label{adse3:11}
    \bsT^\inv = \Ubig(\exp\bsD)^\inv\Ubig^\inv = \Ubig\exp(-\bsD)\Ubig^\inv,
\end{equation}
where
\begin{equation}
    (\exp\bsD)^\inv = \bbm \exp(-\mbf D) & -\mbf F\exp(-\mbf D) \\
        \bfO & \exp(-\mbf D) \ebm = \bbm \exp(-\mbf D) & -\exp(-\mbf D)\mbf F \\
        \bfO & \exp(-\mbf D) \ebm = \exp(-\bsD).
\end{equation}
Using real decomposition,
\begin{equation}\label{adse3:13}
    \bsT^\inv = \bsUp(\exp\bfde^\times)^\inv\bsUp^\inv = \bsUp\exp(-\bfde^\times)\bsUp^\inv,
\end{equation}
where $\bfde$ is given in (\ref{adse:1a}).  In either representation, the argument of the exponential, $-\bsD$ or $-\bfde$, simply demands that $\phi$ and $r$ be replaced with $-\phi$ and $-r$.  Not surprisingly, the inverse pose transformation of $\Ad(SE(3))$ is geometrically obtained by reversing the principal-axis pose.

}

\vspace*{10pt}

\stepcounter{subsection}
\noindent
{\bf \arabic{section}.\arabic{subsection}\,\, The Jacobian $\bsJ$}%
\vspace*{8pt}

\noindent
To underscore the kinship between $SO(3)$ and $\mbox{Ad}(SE(3))$, we note that the kinematical relation for angular velocity, $\mbs\om = \mbf J\dot{\mbs\phi}$ in $\Re^3$, where $\mbf J$ is the left Jacobian given in (\ref{eq:J}), has a direct counterpart in $\Re^6$ for generalized velocity: $\mbs\upsilon = \boldsymbol{\mathcal J}\dot{\bfxi}$, where $\boldsymbol{\mathcal J}$, the left Jacobian of $\Ad(SE(3))$, is also a coexponential \citep{barfoot_ser17}.  Also paralleling the relationship between $\bfC$ and $\bfJ$, $\bsT = \1 + \bfxi^\x\bsJ = \1 + \bsJ\bfxi^\x$.  There is no analogous counterpart in $SE(3)$.

As in $SO(3)$, we can express the left Jacobian $\bsJ$ in $\Ad(SE(3))$ using the foregoing eigendecomposition, which in this case leads to a computational advantage.  The Jacobian here, echoing $SO(3)$, is given by
\begin{equation}\label{J:1}
    \bsJ = \int_0^1 \bsT^\alpha d\alpha.
\end{equation}
Using complex decomposition once again, we have
\begin{equation}
    \bsT^\alpha = \exp(\alpha\bfxi^\Wdg) = \Ubig\exp(\alpha\bsD)\Ubig^\inv.
\end{equation}
We observe that $\bsD$ may be broken up by a Jordan-Chevalley decomposition, i.e.,
\begin{equation}
    \bsD = \bsDe + \bsF,
\end{equation}
where
\begin{equation}
    \bsDe = \bbm \bfD & \bfO \\ \bfO & \bfD \ebm, \qquad \bsF = \bbm \bfO & \bfF \\ \bfO & \bfO \ebm
\end{equation}
and $\bsF$ is nilpotent of degree 2.  The diagonal matrices $\bsDe$ and $\bsF$ of course commute in multiplication.  Hence
\begin{multline}
    \exp(\alpha\bsD) = \exp[\alpha(\bsDe + \bsF)] = \exp(\alpha\bsDe)\exp(\alpha\bsF)
    = \exp(\alpha\bsDe)(\1 + \alpha\bsF) \\ = \bbm \exp(\alpha\bfD) & \alpha\exp(\alpha\bfD)\bfF \\ \bfO & \exp(\alpha\bfD) \ebm.
\end{multline}
Now the integral of $\exp(\alpha\bfD)$ is available from (\ref{eq:intD}) and
\begin{equation}
    \int_0^1 \alpha\exp(\alpha\bfD)d\alpha\mbf F = \text{diag}[\,0, irf(\phi), -irf(-\phi)\,] = \mbf N,
\end{equation}
where
\begin{equation}
  f(\phi) = -\frac{(i\phi - 1)e^{i\phi} + 1}{\phi^2}.
\end{equation}
We can then write
\begin{equation}
    \int_0^1 \exp(\alpha\bsD)d\alpha = \bbm \mbf M & \mbf N \\ \bfO & \mbf M \ebm,
\end{equation}
where $\mbf M$ is given by (\ref{eq:intD}).  Thus, we eventually obtain
\begin{equation}
    \bsJ = \bbm \bfJ & \mbf W\bfU^H\bfJ - \bfJ\mbf W\bfU^H + \bfU\mbf N\bfU^H \\
        \bfO & \bfJ \ebm.
\end{equation}
We can also proceed, however, using real decomposition.

From (\ref{eig:004}),
\begin{equation}\label{J:2}
    \bsT^\alpha = \bsUp\exp(\alpha\bfde^\x)\bsUp^{-1}.
\end{equation}
Let us decompose $\bfde^\x$ as
\begin{equation}
    \bfde^\x = \bsd^\x + \mbs\pi^\x,
\end{equation}
where
\begin{equation}
    \bsd = \bbm \0 \\ \mbf d \ebm = \bbm \0 \\ \phi\1_1 \ebm, \qquad
    \mbs\pi = \bbm r\1_1 \\ \0 \ebm.
\end{equation}
Note that $\bsd^\x$ and $\mbs\pi^\x$ commute in multiplication and that $\mbs\pi^\x$ is nilpotent of degree 2.  Hence
\begin{equation}
    \exp(\alpha\bfde^\x) = \exp(\alpha\bsd^\x)(\1 + \alpha\mbs\pi^\x)
    = \bbm \exp(\alpha\phi\1_1^\x) & \alpha r\1_1^\x\exp(\alpha\phi\1_1^\x) \\
    \bfO & \exp(\alpha\phi\1_1^\x) \ebm.
\end{equation}
Upon integration, the diagonal blocks yield (\ref{eq:intd}) and the off-diagonal block becomes
\begin{multline}
    \int_0^1 \alpha r\1_1^\x\exp(\alpha\phi\1_1^\x)d\alpha
        = -\frac{r}{\phi}\1_1^\x\1_1^\x\exp(\phi\1_1^\x) + \frac{r}{\phi^2}\1_1^\x\left[\exp(\phi\1_1^\x) - \1\right]\\
        = -r\frac{1 - \phi\sin\phi - \cos\phi}{\phi^2}\1_1^\x + r\frac{\sin\phi - \phi\cos\phi}{\phi^2}\1_1^\x\1_1^\x = \mbf K.
\end{multline}
Using (\ref{J:2}) in (\ref{J:1}), we arrive at
\begin{equation}
    \bsJ = \bbm \mbf J & \phi^\inv(\mbf a^\x\mbs\rho)^\x\mbf J - \phi^\inv\mbf J(\mbf a^\x\mbs\rho)^\x + \bfUp\mbf K\bfUp^T \\ \bfO & \mbf J \ebm,
\end{equation}
where
\begin{equation}
    \bfUp\mbf K\bfUp^T = -r\frac{1 - \phi\sin\phi - \cos\phi}{\phi^2}\mbf a^\x + r\frac{\sin\phi - \phi\cos\phi}{\phi^2}\mbf a^\x\mbf a^\x.
\end{equation}
This may be compared to the more involved result \citep{barfoot_ser17},
\begin{multline}
  \bsJ = \1 + \frac{4 - \phi\sin\phi - 4\cos\phi}{2\phi^2}\bfxi^\x
    + \frac{4\phi - 5\sin\phi + \phi\cos\phi}{2\phi^3}{\bfxi^\x}^2 \\
    + \frac{2 - \phi\sin\phi - 2\cos\phi}{2\phi^4}{\bfxi^\x}^3
    + \frac{2\phi - 3\sin\phi + \phi\cos\phi}{2\phi^5}{\bfxi^\x}^4
\end{multline}
obtained by a polynomial expansion for $\bsT$.